\documentclass[letterpaper, 10 pt, conference]{ieeeconf}  % Comment this line out if you need a4paper

\IEEEoverridecommandlockouts                              % This command is only needed if 
\usepackage{graphicx} % for pdf, bitmapped graphics files

\usepackage{caption}
\usepackage{subcaption}
\usepackage{url}
\usepackage{hyperref}

\title{\LARGE \bf
Opportunities and Limitations of Mixed Reality Holograms in Industrial Robotics*
}

\author{Michael Filipenko$^{1}$, Andreas Angerer$^{2}$, Alwin Hoffmann$^{1}$, Wolfgang Reif$^{1}$% <-this % stops a space
\thanks{*This work was carried out in the Innovation Lab for Collaborative Robotics at the University of Augsburg and was supported by the Centre Digitisation.Bavaria (Zentrum Digitalisierung.Bayern, ZD.B).}% <-this % stops a space
\thanks{$^{1}$Michael Filipenko, Alwin Hoffmann, and Wolfgang Reif are with the Institute for Software and Systems Engineering,
        University of Augsburg, D-86135 Augsburg, Germany
        {\tt\small filipenko@isse.de}}%
\thanks{$^{2}$Andreas Angerer is with XITASO GmbH, D-86153 Augsburg, Germany
        {\tt\small andreas.angerer@xitaso.com}}%
}

\begin{document}

\maketitle
\thispagestyle{empty}
\pagestyle{empty}

%%%%%%%%%%%%%%%%%%%%%%%%%%%%%%%%%%%%%%%%%%%%%%%%%%%%%%%%%%%%%%%%%%%%%%%%%%%%%%%%
\begin{abstract}
This paper introduces two case studies combining the field of industrial robotics with Mixed Reality (MR).
The goal of those case studies is to get a better understanding of how MR can be useful and what are the limitations. 
The first case study describes an approach to visualize the \textit{digital twin} of a robot arm. 
The second case study aims at facilitating the commissioning of industrial robots. 
Furthermore, this paper reports the experiences gained by implementing those two scenarios and discusses the limitations.

\end{abstract}

%%%%%%%%%%%%%%%%%%%%%%%%%%%%%%%%%%%%%%%%%%%%%%%%%%%%%%%%%%%%%%%%%%%%%%%%%%%%%%%%
\section{INTRODUCTION} \label{sec:introduction}

Virtual Reality (VR) and Augmented Reality (AR) are technologies which are widely used nowadays~\cite{sanna2016survey}. 
In theme parks, one can ride a roller coaster while wearing a VR headset in order to experience the ride in a completely different world. 
AR applications are common for smartphones, e.g., to place virtual furnitures in one's flat. 
Furthermore, augmented reality has found its way into industry. Applications are, e.g., remote maintenance~\cite{masoni2017supporting} or quality assurance~\cite{frigo2016augmented}.

Milgram~\cite{milgram1995augmented} has defined the term \emph{Mixed Reality} (MR) for any application which is inside a virtuality continuum between the real world and a virtual one. 
%Hence, Mixed Reality describes any application which uses aspects of the real and the virtual world. 
Although introduced in 1995, the term Mixed Reality was not used for some time.
Microsoft rediscovered the term in 2016 again when introducing its HoloLens, a head-mounted see-through display device which is able to perform a spatial mapping and can display 3D holograms precisely in its environment~\cite{MixedReality}. 

This paper introduces two use cases how to make use of MR for industrial robotics (cf. Sect.~\ref{sec:usecases}). The goal was to study the possible advantages of MR technologies to facilitate the commissioning and usage of industrial robotics. Especially 3D holograms allow for displaying a robot's internal state, i.e. the encoded knowledge of the robot's environment as well as informations given by sensors. The case studies have been implemented using the Microsoft HoloLens (cf. Sect.~\ref{sec:implementation}). The insights we gained during our work as well as limitations we discovered are explained in Sect.~\ref{sec:limitations}. Sect.~\ref{sec:conclusions} gives a conclusion.

%%%%%%%%%%%%%%%%%%%%%%%%%%%%%%%%%%%%%%%%%%%%%%%%%%%%%%%%%%%%%%%%%%%%%%%%%%%%%%%%
\section{USE CASES} \label{sec:usecases}

To study opportunities and limitations regarding the use of Mixed Reality in industrial robotics, we chose two case studies we named "Digital Twin Diagnosis" (case study A) and "Mixed Commissioning" (case study B). 
The following sections present the ideas behind those two case studies.

\begin{figure}
	\centering
	\begin{subfigure}[t]{0.48\linewidth}
		\centering
		\includegraphics[width=0.85\textwidth]{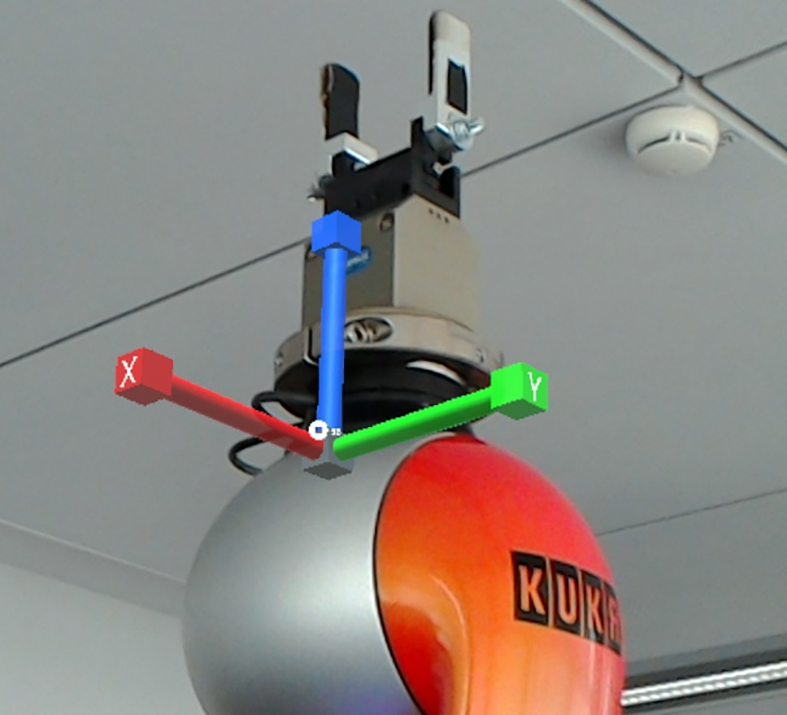}
		\caption{Flange coordinate system displayed as pose hologram.}
		\label{fig:flangeframe}
	\end{subfigure}
	\hfill
	\begin{subfigure}[t]{0.48\linewidth}
		\centering
		\includegraphics[width=0.85\textwidth]{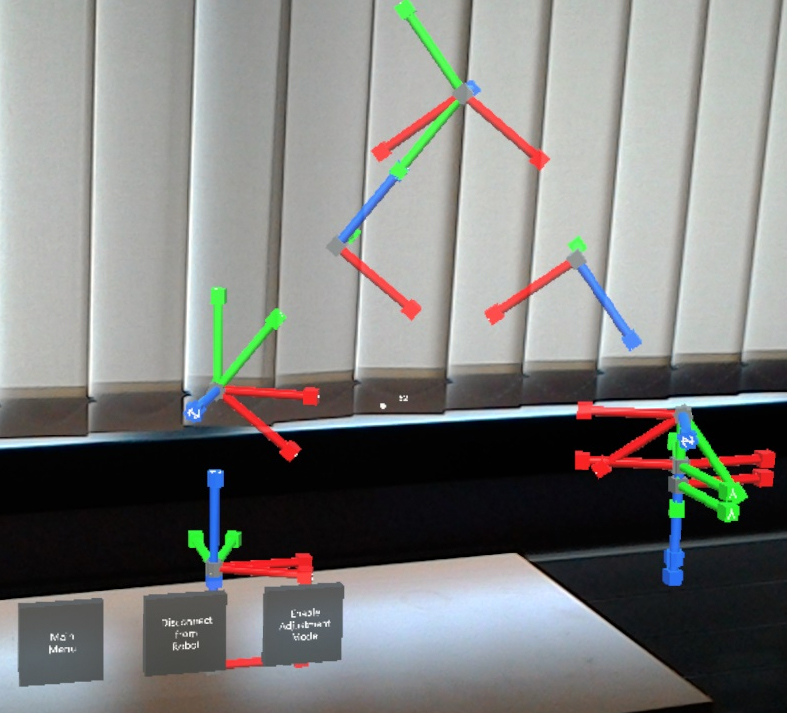}
		\caption{Intermediate points of a motion shown as pose holograms.\newline}
		\label{fig:movementframes}
	\end{subfigure}

	\begin{subfigure}[b]{0.48\linewidth}
		\centering
		\includegraphics[width=0.85\textwidth]{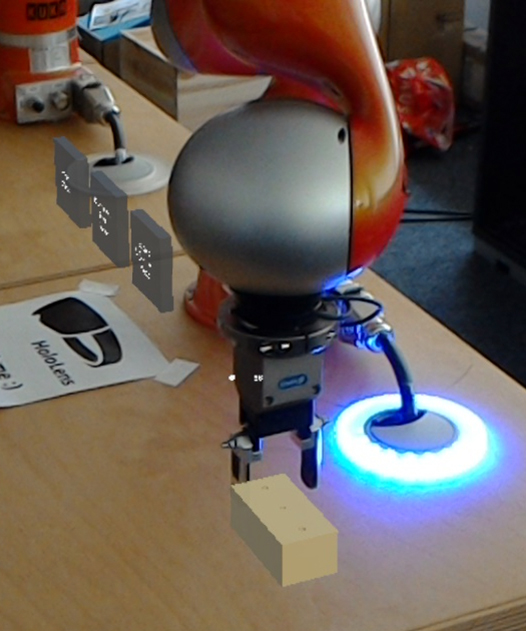}
		\caption{Hologram of a workpiece.}
		\label{fig:workpiece}
	\end{subfigure}
	\hfill
	\begin{subfigure}[b]{0.48\linewidth}
		\centering
		\includegraphics[width=0.85\textwidth]{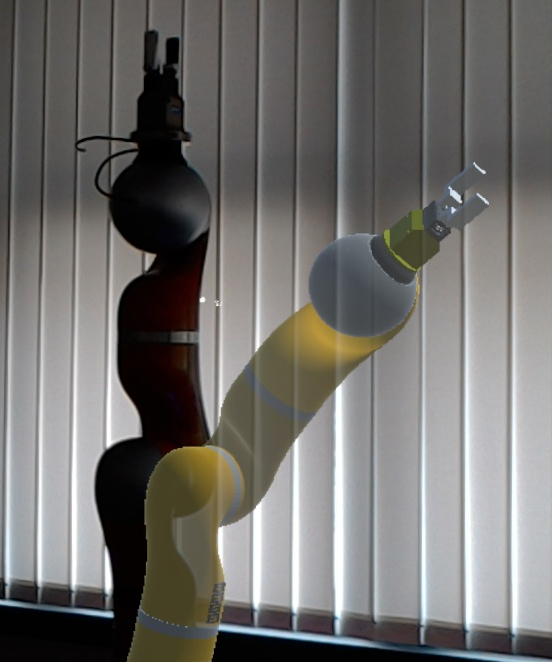}
		\caption{Motion forecast hologram.}
		\label{fig:forecast}
	\end{subfigure}
	\caption{Example holograms for diagnosis purposes.}
	\label{fig:holograms}
\end{figure}

%TODO do we have some image showing sensor data?
%TODO the figures (in part. Fig.~\ref{fig:workpiece}) indicate that this is rather augmented than mixed reality, right? Should we adapt title and text, then?
\subsection{Digital Twin Diagnosis}
This case study aims at using holograms as visualization of the \emph{digital twin} of a robot system inside its environment. 
We define the term digital twin here as the digital information the robot system can access about itself and its environment.
This information is usually determined by a combination of encoded knowledge (e.g. about the robot structure or the shapes of workpieces) and sensed information (e.g. the joint angles according to encoders, locations of workpieces according to cameras). When presented in form of a hologram, this information can easily be used to detect problems during operation and safely diagnose their origin:
By overlays on the environment, gaps between the physical world and its digital twin (in the above sense) can easily be detected by human operators.
Fig.~\ref{fig:flangeframe} to Fig.~\ref{fig:workpiece} show some example holograms that can be useful for diagnosis purposes.

Furthermore, the safety of operators and the environment can be increased by including holographic forecasts of the system's actions (cf. Fig.~\ref{fig:forecast})
In this way, robot movements that might lead to collisions or other inappropriate actions can easily be identified, which can greatly assist diagnosis and re-commissioning.
Finally, the Mixed Reality approach is beneficial for remote diagnosis as well -- 
% @ Andi: wie ist der auskommentierte Teil genau gemeint? 
% even if a remote assistant does not have appropriate hardware at hand, he can get a better impression of the situation by watching the augmented video stream on a regular monitor.
% Alternativvorschlag:
the visualization created for the onsite operator could easily be used by an expert on a distant location. 
Of course there won't be an overlay with the real robot, but the digital twin should provide all the information needed to help solving the problem.
Even if a remote expert does not have an appropriate device at hand, the augmented video stream can be helpful when watched on a regular monitor.

\subsection{Mixed Commissioning}
\begin{figure}
	\centering
	\includegraphics[width=0.9\linewidth]{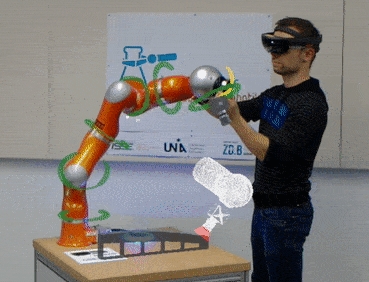}
	\caption{Holograms for commissioning support.}
	\label{fig:commissioning}
\end{figure}

The second case study is an extension of the Digital Twin Diagnosis case study presented above. 
Here we envision Mixed Reality support for commissioning of robot systems, in particular robot programming or optimization of movement and tool parameters.
By using holograms for (future) end-effector positions, workpiece positions or planned movement trajectories, new possibilities for efficient teach-in of robot actions arise:
During on-line teach-in or hand guiding of a robot arm, an indication of the joint limits can be helpful (cf. Fig.~\ref{fig:commissioning}).
For certain robot tools or sensing devices, it is beneficial to augment information about the devices' work area.
For example, a camera system that is used for quality inspection should be moved in a way that respects the camera field of view or its focus area.
The lower part of Fig.~\ref{fig:commissioning} indicates the field of view of a simulated camera as a hologram of a red cone.

%%%%%%%%%%%%%%%%%%%%%%%%%%%%%%%%%%%%%%%%%%%%%%%%%%%%%%%%%%%%%%%%%%%%%%%%%%%%%%%%
\section{IMPLEMENTATION} \label{sec:implementation}

\begin{figure}
	\centering
	\includegraphics[width=\linewidth]{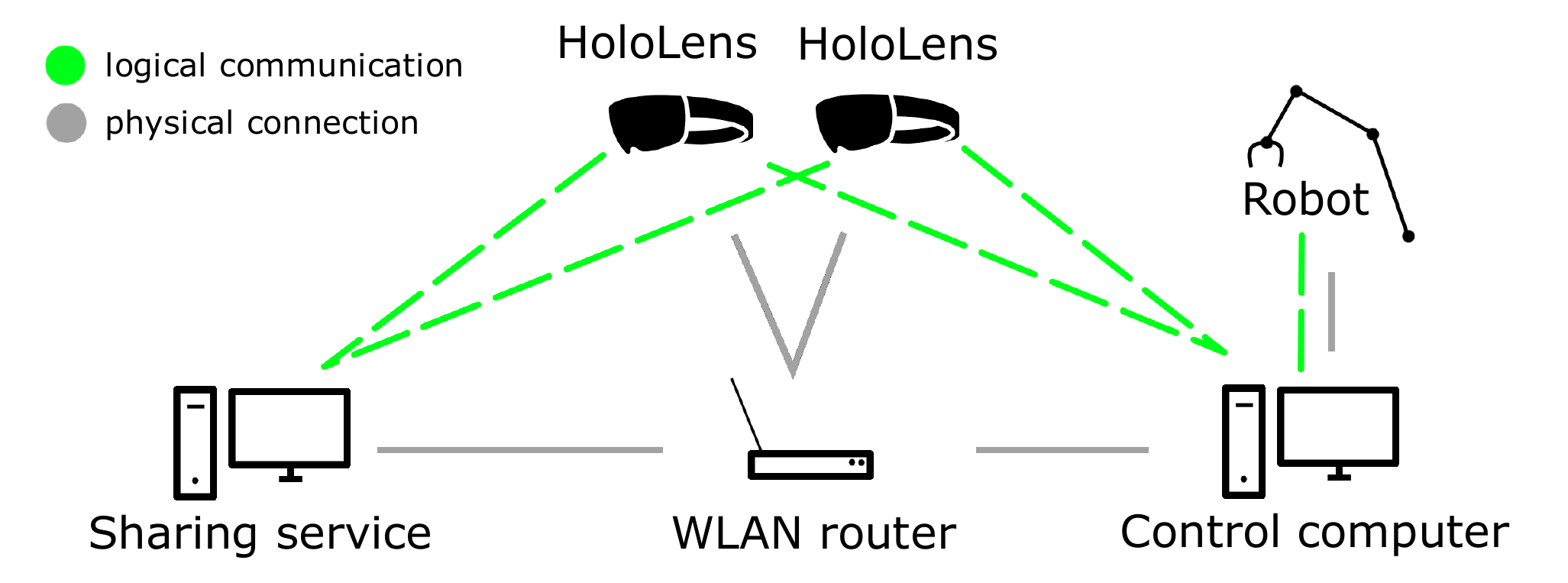}
	\caption{Mixed Reality prototype overview.}
	\label{fig:architecture}
\end{figure}

To evaluate the presented case studies, a prototypical Mixed Reality system was built as illustrated in Fig.~\ref{fig:architecture}.
The Microsoft HoloLens (first generation) was used as a headmounted Mixed Reality device.
As indicated in the figure, the system was built to support multiple devices from the start in order to foster collaborative use.
Other users can share the mixed reality video stream on a standard monitor or their own HoloLens as well.
In the latter case, they can interact with the system just as the main user in a collaborative fashion.
A standard WiFi access point was used as a dedicated network infrastructure.
One standard desktop computer hosted a Sharing Service infrastructure that has been developed for this work.
A second desktop computer was responsible for controlling the industrial robot used in this work, a KUKA LWR-4~\cite{Bischoff2010}.
Robot programming and control was done using the Robotics API~\cite{RoboticsAPI} connected to the LWR's KRC2lr controller.

To build the application infrastructure, a combination of Microsoft's Mixed Reality Toolkit~\cite{HoloToolkit} (formerly called HoloToolkit), the Unity framework~\cite{Unity} and Vuforia~\cite{Vuforia} was used.
The HoloLens depth camera was used via HoloToolkit for depth mapping, whereas the RGB camera is used for additional 2D marker tracking based on Vuforia's object recognition functionality.
A key finding was that depth mapping alone is not robust enough in dynamic environments, therefore the marker tracking couldn't be used only at the start for an initial space calibration but rather 
had to stay active the whole time. To synchronize the content between multiple
HoloLens devices the sharing service of the HoloToolkit was used. The current robot data
was multicasted into the network via UDP and could be consumed by any member of
the network.  
When it comes to user input, mainly the gesture recognition of the HoloLens was used.

Due to the limited amount of gestures the HoloLens provides (AirTap and Drag) while simultaneously making the whole environment to a potential user interface, the interaction design was given a lot of thought. 
A first straightforward concept -- based on a special 2D window
floating next to the robot and providing holographic buttons to hide or show
each hologram in the scene -- showed deficiencies:
this way of interaction with the holograms was pretty
tedious because the operator had to turn his head away from the robot every time he would like to
change something about the hologram appearance - therefore creating an actual
separation between virtual content and real objects. 

\begin{figure}
	\centering
	\includegraphics[width=0.85\linewidth]{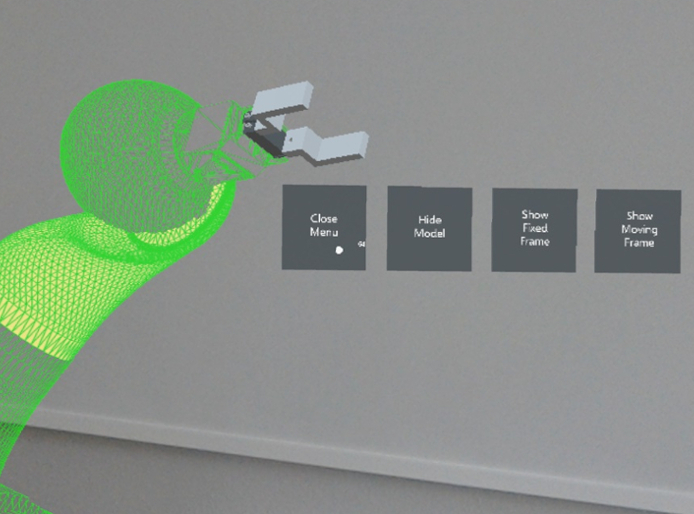}
	\caption{Interaction example to manipulate part visibility.}
	\label{fig:interaction}
\end{figure}

To merge the virtual and real world, the refined approach shown in Fig.~\ref{fig:interaction} was used.
It fosters an interaction model based on a combination of "classical" holographic buttons and all other available holograms. 
The buttons were used to trigger specific events (e.g. hiding or showing parts of the robot) and where designed in a way that they would always face directly to the user. 
The other holograms (regarding the robot itself in this case) were used as a kind of contextual menu. 
The basic idea behind it can be formulated as follows: 
An operator should be able to gaze at a hologram from any angle and, by activating it via AirTap, he should get all currently available information regarding the specific part he had chosen.

The goal was to mimic the interaction humans have with real objects as
best as possible and as a result blur the border between virtual and real world.
This contextual menu has three main benefits: 
The first one was to reduce the amount of information
shown to the user simultaneously. 
The second one was to eliminate the problem of the
operator turning away when trying to change something about the scene. 
And the third one was to make the interaction possible from any angle and
therefore allow the user to move freely.
However, for some global actions, also voice commands were employed.
For that, we used the speech recognition feature provided by Microsoft's Mixed Reality Toolkit~\cite{HoloToolkit}, which allows to specify arbitrary keywords in English language.

%%%%%%%%%%%%%%%%%%%%%%%%%%%%%%%%%%%%%%%%%%%%%%%%%%%%%%%%%%%%%%%%%%%%%%%%%%%%%%%%
\section{INSIGHTS \& LIMITATIONS} \label{sec:limitations}

This section will explain insights gained during our work as well as limitations we encountered and outline possible mitigations\footnote{Because the experiments have been made with the Microsoft HoloLens (first generation), the limitations sometimes refer to this specific device.}.

\subsection{Limited field of view}
\begin{figure}
	\centering
	\includegraphics[width=0.85\linewidth]{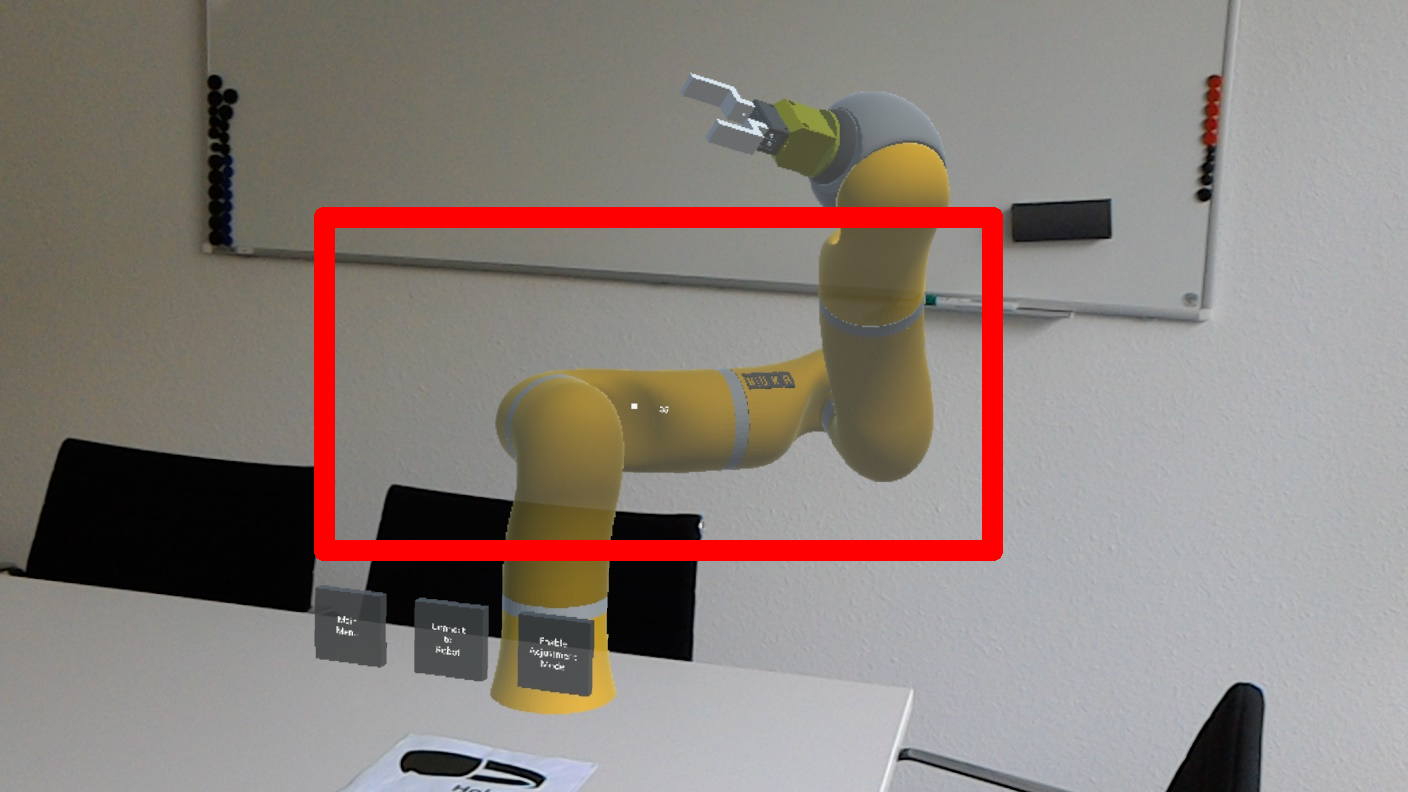}
	\caption{The field of view for holograms of the Microsoft HoloLens is  marked (in red).}
	\label{fig:hololensFOV}
\end{figure}

One of the biggest and most obvious limitations of the HoloLens is the limited holographic field of view which showed to be problematic in our experiments.
Fig.~\ref{fig:hololensFOV} shows the approximate relative holographic field of view compared to the complete field of vision of an operator.
This limitation forces operators to artificially limit their head movements in order not to lose important holographic information, which in turn makes it even more important to design the information shown to the operator very carefully. More recent devices such as the Magic Leap or the HoloLens 2 have a wider field of view which probably will mitigate this issue.

\subsection{UI/UX design}
The UI/UX design of an AR/MR system is non-trivial.
The presented prototype uses spatial gestures in most cases and places dedicated holograms in the environment.
For an optimal user experience, the holograms' spatial distance to the operator is a crucial factor.
Microsoft's official suggestion is to keep distance between 80cm and 200cm~\cite{hololens-doc}.
As our use cases involve working with a robot arm in human arm's length, we regularly had a distance of about 40cm between operator and holograms.
We did not experience specific problems here. However the user interface had to
be designed for a three-dimensional space and for a very limited field of view,
which made it quite challenging to decide what information to show and in what way.

\subsection{Multi-modal interaction}
As mentioned before, voice commands are a viable addition to gesture interaction.
In particular, voice commands provide a better experience if interaction is not naturally linked to real world or holographic scene objects.
For robust recognition, a careful design of clearly distinguishable voice commands showed to be vital.
However, an important open question for practical usage is the robustness of speech recognition in industrial environments.
HoloLens 2 promises improvements regarding speech recognition \cite{HoloLens95:online}, which could mitigate potential issues.
An other issue that occurred when multiple users were using a HoloLens in the same space was that sometimes a voice command from one user was picked up by the wrong device. 
Particularly in crowded environments, this could become an issue.

\subsection{Comprehensibility for others}
Systems based on interaction using eye-glass devices show information to the immediate operator at first instant.
For others, the interaction is not directly conceivable due to missing information, which hinders collaboration.
An easy mitigation can be achieved by displaying the operator's visual stream on a separate monitor.
However, the latency has to be as low as possible because otherwise the collaboration becomes tedious and annoying.
In this work, the latency using the HoloLens and the local network setup proofed to be sufficient for seamless collaboration.
Nevertheless, the low resolution of the HoloLens video stream is problematic, as hologram texts have been barely readable for others.
%(color separation) yes it is an issue, but I would only mention it, if we have
%way to much space left, because it isn't really that dramatic.

\subsection{Safety}
Due to its mechanical design, the HoloLens and possible other devices limit an operator's field of vision.
Furthermore, a careful design of truly \emph{mixed} reality behavior showed to be vital, as operators can be easily distracted if real-world objects are hidden by holograms.
Both above mentioned aspects can lead to safety issues in industrial environments.
Furthermore, it can be difficult to combine mixed reality devices with industrial helmets which hinders safety as well.

\subsection{Tracking robustness in dynamic environments}
The presented experiments revealed that the current spatial mapping of the HoloLens has limited robustness when it comes to changes in the physical environment.
This limits its applicability in human-robot collaboration scenarios that usually involve manipulation of mobile objects in the environment.
Marker-based tracking is per se less prone to that (as long as markers are visible).
However, fast operator movements were problematic for both localization approaches.

\subsection{Performance issues}
The embedded processing power of the HoloLens limits the complexity of holographic scenes.
Generally speaking it has almost the same limits as other devices such as mobile phones.
According to available information, scenes with up to 100k polygons can be handled\footnote{\url{https://twitter.com/hololens/status/756247350633455620}}, but shader optimizations might be necessary.
% Otherwise here is a recommendation page from microsoft, which provides infos
% to that topic, but no concrete numbers. https://docs.microsoft.com/de-de/windows/mixed-reality/asset-creation-process
In our experiments, the use of marker tracking lead to significant performance drops:
Average performance dropped from 60fps to 40fps, leading to instability of the hologram positions -- especially noticeable during fast movements.

\subsection{Imprecise hologram alignment}
Due to various reasons, the alignment of holograms with real-world objects is hard to achieve~\cite{liu2018technical}.
Reasons include e.g. imperfect localization, incorrect recognition of physical object sizes or incorrect scale of hologram models.

%%%%%%%%%%%%%%%%%%%%%%%%%%%%%%%%%%%%%%%%%%%%%%%%%%%%%%%%%%%%%%%%%%%%%%%%%%%%%%%%
\section{CONCLUSIONS} \label{sec:conclusions}

In this paper, we have presented two case studies how mixed reality can be useful in the field of robotics. 
The first one was about visualizing the \textit{digital twin} of a robot arm and analyzing what benefits this new way to represent robot data could have.
The two main aspects we discovered were its usage to diagnose problems faster and also provide a safer way of interacting with a robot arm, for example by forecasting its movements.
The second case study showed a method how mixed reality could be used in commissioning. 
Important data about the robot joint values or workpieces can be shown during teach-in or programming and therefore could help to achieve the desired goals quicker or with less mistakes.

We have implemented an application featuring both case studies in order to determine the advantages of MR. 
Thereby we encountered some limitations considering the used mixed reality hardware, which in this case was a HoloLens (first generation). 
The biggest issues are the limited field-of-view and the limitations in processing power due to its mobility.
These limitations will probably be solved by future generations of MR device such as the Magic Leap or the HoloLens 2. Additionally, they will offer new interaction models and an advance gesture tracking.
However, a conceptional challenge will remain: designing a good MR user interface around a robot and its environment which feels natural and intuitive is anything but a trivial task and therefore offers many research possibilities in the future.

\addtolength{\textheight}{-0cm}   % This command serves to balance the column lengths
                                  % on the last page of the document manually. It shortens
                                  % the textheight of the last page by a suitable amount.
                                  % This command does not take effect until the next page
                                  % so it should come on the page before the last. Make
                                  % sure that you do not shorten the textheight too much.

%%%%%%%%%%%%%%%%%%%%%%%%%%%%%%%%%%%%%%%%%%%%%%%%%%%%%%%%%%%%%%%%%%%%%%%%%%%%%%%%
\section*{APPENDIX}

A video\footnote{\url{https://youtu.be/RnTraZt1lEk}} can be found on the YouTube channel of the Institute for Software and
Systems Engineering to get a better understanding of the described concepts.

%%%%%%%%%%%%%%%%%%%%%%%%%%%%%%%%%%%%%%%%%%%%%%%%%%%%%%%%%%%%%%%%%%%%%%%%%%%%%%%%
\section*{ACKNOWLEDGMENT}

The authors thank Christian H\"ofle and Jan G\"untner, both XITASO GmbH, for their continuous and helpful support during this work. 
Furthermore, we thank the Centre Digitisation.Bavaria (Zentrum Digitalisierung.Bayern, ZD.B) for their financial and organisatorial support.

%%%%%%%%%%%%%%%%%%%%%%%%%%%%%%%%%%%%%%%%%%%%%%%%%%%%%%%%%%%%%%%%%%%%%%%%%%%%%%%%

\bibliographystyle{plain}
{\small
	\bibliography{paper}}
\end{document}